\documentclass[english]{article}

\usepackage[preprint,nonatbib]{neurips_2023}

\usepackage[utf8]{inputenc} 
\usepackage[T1]{fontenc}    
\usepackage[colorlinks]{hyperref}       
\usepackage{url}            
\usepackage{booktabs}       
\usepackage{amsfonts}       
\usepackage{nicefrac}       
\usepackage{microtype}      

\usepackage{amsmath}
\usepackage{amssymb}
\usepackage{amsthm}
\usepackage{mathtools}
\usepackage{graphicx}
\usepackage[table]{xcolor}
\usepackage[english]{babel}
\usepackage[capitalize]{cleveref}

\usepackage{algorithm}
\usepackage{fontawesome5}
\usepackage{algpseudocode}

\usepackage[most]{tcolorbox}
\usepackage{accents}
\usepackage{enumitem}
\usepackage{wrapfig}
\usepackage{booktabs}
\usepackage{colortbl}
\usepackage{array}
\usepackage{multirow}
\usepackage[numbers]{natbib}
\usepackage{tabularx}   

\usepackage{caption}
\usepackage{subcaption}

\usepackage{MnSymbol}

\newcommand\munderbar[1]{%
  \underaccent{\bar}{#1}}

\newtcolorbox{myquote}[1][]{%
    title=#1,
    colback=green!4,
    colframe=black!10,
    notitle,
    sharp corners,
    borderline west={2pt}{0pt}{red!80!black},
    enhanced,
    breakable,
    }

\newtcolorbox{tabbedbox}[1]{
  enhanced,
  breakable,
  title={#1},
  fonttitle=\bfseries\large\sffamily,
  attach boxed title to top left={xshift=1cm, yshift=-\tcboxedtitleheight/2},
  colbacktitle=blue!75!green!75!black,
  coltitle=white,
  colframe=blue!75!green!50!black,
  colback=blue!5!green!5!white,
  boxed title style={
    sharp corners, 
    size=small,
    colback=blue!75!green!75!black
  },
  boxrule=0.5mm,
  left=5mm,
  top=15pt
}

\title{PLoP: Precise LoRA Placement for Efficient Finetuning of Large Models
\author{%
  Soufiane Hayou\thanks{Corresponding author: \texttt{hayou@berkeley.edu}}\\
  Simons Institute\\
  UC Berkeley \\
  \And
  Nikhil Ghosh \\
  Flatiron Institute\\
  \And
  Bin Yu\\
  Dept of Statistics\\
  UC Berkeley\\}
}
\author{}
\date{}

\newcommand{\Zin}{z_{in}}
\newcommand{\ZinT}{\Tilde{z}_{in}}
\newcommand{\Zout}{z_{out}}
\newcommand{\bigO}{\mathcal{O}}
\newcommand{\reals}{\mathbb{R}}
\newcommand{\normal}{\mathcal{N}}
\newcommand{\E}{\mathbb{E}}

\newcommand{\Sign}{\mathcal{S}}

\newcommand{\data}{\mathcal{D}}

\newcommand{\method}{\texttt{PLoP}}

\definecolor{headercolor}{RGB}{52, 101, 164}
\definecolor{rowcolor1}{RGB}{233, 243, 255}
\definecolor{rowcolor2}{RGB}{255, 255, 255}

\newtheorem{thm}{Theorem}

\newtheorem{definition}{Definition}

\definecolor{niceblue}{rgb}{0.10, 0.14, 0.76} 

\definecolor{nicered}{rgb}{0.70, 0.0, 0.0}

\AtBeginDocument{%
\hypersetup{
    citecolor=niceblue,
    linkcolor=red,   
    urlcolor=niceblue,
    linktoc=none}
}

\begin{document}

\maketitle
\vspace{-4em}
\begin{figure}[h]
    \centering
    \includegraphics[width=0.4\linewidth]{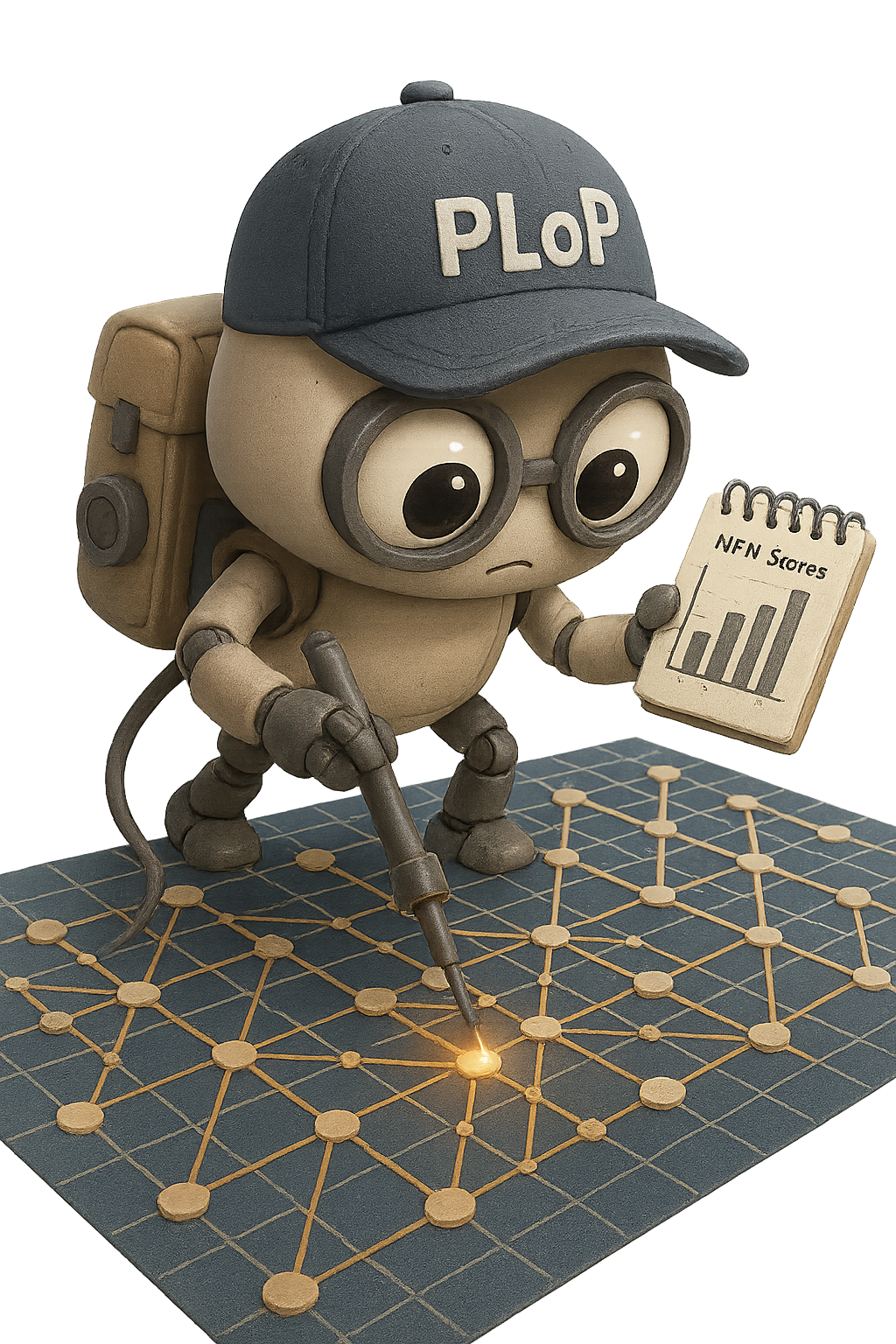}
\end{figure}

\begin{abstract}
Low-Rank Adaptation (LoRA) is a widely used finetuning method for large models. Its small memory footprint allows practitioners to adapt large models to specific tasks at a fraction of the cost of full finetuning.  Different modifications have been proposed to enhance its efficiency by, for example, setting the learning rate, the rank, and the initialization. Another improvement axis is adapter placement strategy: when using LoRA, practitioners usually pick \emph{module types} to adapt with LoRA, such as Query and Key modules. Few works have studied the problem of adapter placement, with nonconclusive results: original LoRA paper suggested placing adapters in attention modules, while other works suggested placing them in the MLP modules. Through an intuitive theoretical analysis, we introduce \method~({\bf P}recise {\bf Lo}RA {\bf P}lacement), a lightweight method that allows automatic identification of module types where LoRA adapters should be placed, given a pretrained model and a finetuning task. We demonstrate that \method~consistently outperforms, and in the worst case competes, with commonly used placement strategies through comprehensive experiments on supervised finetuning and reinforcement learning for reasoning.
\end{abstract}

\section{Introduction}\label{sec:intro}
Low-Rank Adaptation (LoRA) is a widely adopted parameter-efficient fine-tuning (PEFT) methods for large language and vision models. Introduced by \cite{hu2022lora}, LoRA significantly reduces the computational and memory requirements of finetuning by freezing the pretrained model weights and inserting low-rank matrices into the model. This approach has enabled the adaptation of production-scale models on limited hardware resources while achieving performance comparable to full finetuning.

\paragraph{LoRA improvements.} Several works have considered improving LoRA performance by e.g. using different learning rates for LoRA modules \citep{hayou2024loraplus}, using normalized updates \citep{liu2024dora}, setting adaptive LoRA rank \citep{kim2024rankadaptive, lu2025adaptiverankreducedforgetting}, improving initialization \citep{hayou2024lorainit}, and many other variants, e.g. \citep{zhang2023adaloraadaptivebudgetallocation, dettmers2023qlora, kopiczko2024veravectorbasedrandommatrix, zhang2023lorafamemoryefficientlowrankadaptation, tian2024hydraloraasymmetricloraarchitecture,jiang2024morahighrankupdatingparameterefficient}.

A critical aspect of LoRA is module selection - deciding which specific  components of the model should receive the low-rank adaptation. In practice, instead of selecting individual modules, one selects module types such as ``q\_proj'' (Query modules), ``v\_proj'' (Value modules), etc. In \cite{hu2022lora}, the authors suggested that inserting LoRA in attention modules (Query, Key, and Value) generally yields the best performance among other possible placements. However, in a recent note \citep{fomenko2024note}, the same authors further explained the difficulty encountered in LoRA adapter placement, and mentioned that optimal placement depends on pretrained model and the finetuning task. Another work \cite{he2021towards} found that for some models, placing LoRA adapters in MLP modules gives better performance. Faced with this confusion, practitioners generally follow one of these guidelines or insert LoRA adapters in all modules which comes at a higher finetuning cost. Therefore, it is natural to ask:
\begin{center}
    \textit{Given a model and a task, how can we \underline{select target module types} for LoRA at a \underline{reasonable cost}?}
\end{center}
\paragraph{Memory footprint of LoRA.}
In practice, LoRA is used to finetune large models with relatively low cost. Consider Llama3.2-3B \citep{2024llama3herdmodels}, processing sequences of 2048 tokens with a batch size of 8. With full finetuning, the memory requirements are substantial. The model parameters require 12GB in float32, while the Adam optimizer states add another 24GB. The activations for a single forward pass consume approximately 48GB of memory. This brings the total memory requirement to approximately 84GB necessitating high-end GPUs. This becomes more problematic with larger, production-scale models. With LoRA, the computational cost changes dramatically. Using rank-16 adapters on query and value modules introduces only 10 million trainable parameters (0.33\% of the model). Notably, since gradients are only computed for the adapter weights, the memory overhead for gradient computation is reduced by over 99\%. This enables finetuning on a single 24GB GPU with the same batch size and sequence length. These low memory footprint is what makes LoRA attractive for finetuning. 

\paragraph{Anatomy of a practical module selection method for LoRA finetuning.}
Based on the computational constraints outlined above, any \emph{practical module selection method for LoRA adapter placement must operate within these resource limitations}. We identify three main pillars of a practical method: (i) the method cannot require computing gradients with respect to the full model parameters, as this would defeat the primary purpose of using LoRA, (ii) the selection process should not necessitate multiple forward passes through different model configurations, as this would multiply the already significant activation memory requirements by the number of candidate configurations being evaluated, (iii) the method must avoid storing large intermediate computations or maintaining extensive state across different module evaluations, which would further strain memory resources.  Only methods satisfying these stringent requirements can truly serve practitioners operating in the resource-constrained environments where LoRA provides its greatest value. 

\begin{figure}
    \centering
    \includegraphics[width=0.99\linewidth]{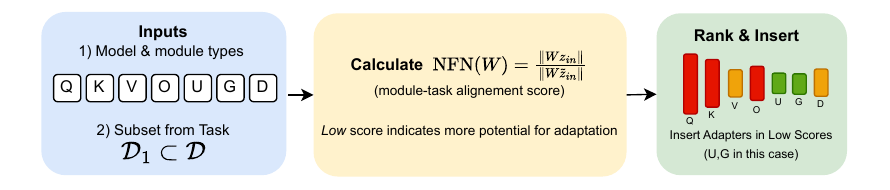}
    \caption{\small{Mechanism of \method. We calculate alignment scores called NFN (Normalized Feature Norms), rank them, and pick module types with the lowest alignment scores for LoRA insertion.}}
    \label{fig:diagram_plop}
\end{figure}

In this paper, we introduce \method~(\textbf{P}recise \textbf{Lo}RA \textbf{P}lacement), a lightweight module  placement method for LoRA based on a specific measure of module-data alignment that can be calculated with few forward passes (no gradients, no extensive  forward passes, and no storage of intermediate calculations), and therefore, it checks all the three points above (see the compute cost paragraph in \cref{sec:methodology} for more details). The mechanism of \method~is described in \cref{fig:diagram_plop}. Specifically, our contributions are as follows: 

\begin{enumerate}
    \item We develop a theoretical framework to study module-data alignment in large neural networks, the core concept behind \method.
    \item Based on our theoretical analysis of module-data alignment, we develop \method, which identifies which module types should be used for LoRA finetuning.
    \item We validate our results with extensive experiments showing the benefits of \method~with LoRA in three post-training scenarios: supervised finetuning for classification, supervised finetuning for text generation, and reinforcement learning for mathematical reasoning. 
\end{enumerate}

The paper is structured as follows. In \cref{sec:theory}, we introduce the main theoretical intuition behind our method. In \Cref{sec:methodology}, we present our method \method~and provide a quantitative and qualitative analysis of our method. In \Cref{sec:exps}, we report empirical results showing the benefit of \method~in two post-training scenarios: supervised finetuning and reinforcement learning. 
\subsection{Related Work}
The effectiveness of LoRA critically depends on the placement of adapter modules. Initially, \citet{hu2022lora} studied the placement of adapters in attention modules, observing strong performance in various NLP tasks. \citet{he2021towards} showed that adapters placed in MLP modules can sometimes outperform attention-based placements. \citet{fomenko2024note} mentioned that optimal adapter placement varies significantly depending on the pretrained model architecture and the downstream task. The authors recommended the following general strategy for adapter placement: start with attention layers, then embeddings, then MLP blocks, and if further capacity is required, raise the LoRA rank.  In machine translation, \citet{gheini2021cross} found that tuning exclusively cross-attention parameters within encoder-decoder Transformers could achieve performance comparable to full-model tuning.

More adaptive approaches include sensitivity-based parameter selection methods. \citet{zhang2024gradient} proposed a gradient-based scoring approach that ranks parameters according to their importance to the task, tuning only the highest scoring subset. Similarly, \citet{he2023sensitivity} developed a sensitivity-aware fine-tuning technique for vision models that dynamically assigns tunable parameters to layers based on local responsiveness. However,  such methods require calculating and storing gradients of the full model which is suboptimal for LoRA finetuning (see discussion above). Another variant of LoRA \cite{zhang2023adaloraadaptivebudgetallocation} introduces modifications to the adapter structure to adaptively distribute capacity between modules. However, our focus in this paper is on module type selection for LoRA. In our experiments, we compare with two baselines: Insertion in attention modules as recommended by \cite{hu2022lora}, and Insertion in MLP modules as recommended by \cite{he2021towards}.




\section{Setup and Theory}\label{sec:theory}
Given a pretrained model and a finetuning task, our goal is to strategically place LoRA adapters in modules that would contribute most significantly to performance. In practice, we usually select module types instead of single weight matrices. For instance, for Llama3 models, we might choose to insert LoRA in Query (``q\_proj") and Key (``k\_proj") modules. 

As we discussed in \cref{sec:intro}, for such a method to be useful, it should first be a lightweight method that operates efficiently in resource-constrained environments and ideally rely on existing computation pipelines. Ideally, the method should rely exclusively on standard forward propagation, as this computational pipeline is already necessary for inference and adds no significant overhead to the existing workflow.

\begin{wrapfigure}{r}{0.3\textwidth}
  \begin{center}
  \vspace{-1em}
    \includegraphics[width=0.29\textwidth]{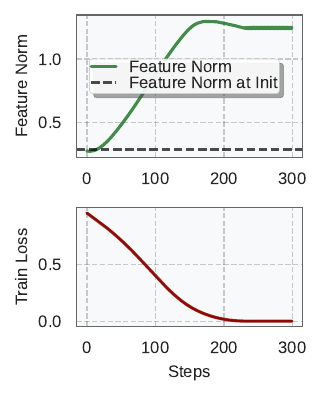}
  \end{center}
  \caption{\small{Illustration of feature norm growth during training. This shows the feature norms $n^{-1}\|W \Zin\|^2$ for a module $W$ in the model ($W \in \reals^{n\times n}$). See \cref{subsec:toy_exps} for details about the model and training.}}
  \label{fig:setup_plot}
  \vspace{-4em}
\end{wrapfigure}

Inspired by this, we investigated the behavior of the activation norms and discovered an interesting phenomenon: \emph{when models are trained on a specific dataset, feature norms of certain modules for inputs from the training data exhibit substantial increases during training, while the same metrics for other modules remain roughly constant}. We show an example of such increase in feature norms in \cref{fig:setup_plot} (see \cref{subsec:toy_exps} for more details).

\paragraph{Mechanisms behind the growth in feature norms.}The reason behind this growth in feature norms for certain modules is non-trivial. The naive explanation to this phenomenon is that with training, weight norms grow for some modules and remain constant constant or decrease for others. However, as we will see in the next analysis, the mechanisms behind this phenomenon are more subtle, and the most important factor is a form of alignment that occurs between module weight and its input.

Specifically, we show that this growth in feature norms in some modules appears primarily as a result of two factors: 1) large-width in neural networks (large embedding dimension), a condition that is generally satisfied in practice, \footnote{From the literature on infinite-width theory, when we take the width to infinity, the training dynamics converge with a rate of roughly $\bigO(n{^{-1/2}})$ \citep{yang2022featurelearninginfinitewidthneural},. In practice, a width of $n \gtrapprox 10^3$ is generally considered large enough for the theoretical predictions to be a good approximation of practice.}, and 2) progressive alignment of modules weights with their respective inputs.

Consider a general neural network of the form 
\begin{equation}\label{eq:model}
\begin{cases}
Y_{in}(x) = W_{in}x,\\
Y_l(x) = \mathcal{F}_l(W_l,Y_{l-1}(x)), \; l\in[L],\\
Y_{out}(x) = W_{out} Y_L(x), 
\end{cases}
\end{equation}

where $x\in\reals^{d}$ is the input, $L\geq1$ is the network depth,
$(\mathcal{F}_{l})_{l\in[L]}$ are mappings that define the layers, $W_{l}\in\reals^{n\times n}$ are the hidden weights, where $n$ is the
network $\emph{width}$, and $W_{in}, W_{out}$ are input and output embedding weights. 

Model \eqref{eq:model} is pretrained on some data mixture $\data$ to minimize some loss function $\ell$ -- the next-token prediction loss in the case of language models. We introduce some notation that will facilitate the presentation of our analysis.
\paragraph{Notation.} Hereafter, $n$ will always denote model width. As $n$ grows, given sequences $c_n \in \reals$ and $d_n \in \reals^+$, we write $c_n = \bigO(d_n)$ to refer to $c_n < \kappa d_n$  for some constant $\kappa > 0$. We write $c_n = \Theta(d_n)$ if we have $\kappa_1 d_n\leq c_n \leq \kappa_2 d_n$ for some $\kappa_1, \kappa_2 >0$. For vector sequences $c_n = (c_n^i)_{1 \leq i \leq k} \in \reals^k$ (for some $k >0$), we write $c_n = \bigO(d_n)$ when $c_n^i = \bigO(d_n^i)$ for all $i \in [k]$, and same holds for other asymptotic notation. Finally, when the sequence $c_n$ is a vector of random variables, asymptotics are defined in the sense of the second moment ($L_2$ norm). For a vector $z\in \reals^n$, we will use the following norms: $\|z\| = \left(\sum_{i=1}^n z_i^2\right)^{1/2}$ (euclidean norm), and $\|z\|_1 = \sum_{i=1}^n |z_i|$ ($\ell_1$ norm).

\paragraph{Intuitive theoretical analysis.} For the sake of tractability, we consider the case where a single weight matrix (module) in the model is trained and other modules are frozen. \footnote{While this is unrealistic, it provides the right intuition behind our methodology, and makes the analysis more tractable.} We further simplify the analysis by assuming that the model is trained in a single datapoint $x$. We later discuss the impact of batch training. The trainable module has the form
$$
z_{out} = W z_{in},
$$
where $z_{in} \in \reals^{n}$ is the input, and $z_{out} \in \reals^{n}$ is the output that we call \emph{feature}, both evaluated at the training datapoint $x$.\footnote{Here we consider that $\Zin$ and $\Zout$ have the same dimension $n$. However, our analysis can be extended to the case where they have different dimensions.} For Transformer models, the module can be for instance a single Query head, a Projection module in some MLP, etc. 

The gradient of the loss with respect to the weight matrix $W$ is given by 
$$
dW = d\Zout \otimes \Zin,
$$
 where $dz_{out} = \nabla_{\Zout}\ell$, the gradient of the loss with respect to feature $\Zout$.

In general, modern LLMs are trained with Adam \citep{kingma2017adammethodstochasticoptimization}, which normalizes gradients. In its momentum-less form, Adam becomes SignSGD \citep{bernstein2018signsgdcompressedoptimisationnonconvex}, which is defined by 
\begin{equation*}
\Sign(dW_{ij}) = 
\begin{cases}
+1 \quad dW_{ij}\geq 0\\
-1 \quad dW_{ij}<0.
\end{cases}
\end{equation*}
SignSGD is a nice simplification of Adam: it captures the property of normalization and allows tractable the theoretical analysis as we will see below. With SignSGD, feature updates\footnote{Feature update is the change of the features $\Zout$ after taking one training step.} are given by

\begin{equation}\label{eq:one_step_signsgd}
\begin{aligned}
W_{t+1}\Zin &= W_t \Zin - \alpha \times \Sign(d\Zout \otimes \Zin) \Zin \\
&= W_t\Zin - \alpha \times \|\Zin\|_1 \, \Sign(d\Zout^t),
\end{aligned}
\end{equation}

where the superscript in $\Zout^t = W_t \Zin$ refers to update step $t$. Note that we do not use such superscript for $\Zin$ since it does not change when we update $W$. 

The key trick used in \cref{eq:one_step_signsgd} is that the sign function $\Sign(.)$ can be expanded across outer product. This is one of the main observations behind the development of $\mu P$ \citep{yang2022featurelearninginfinitewidthneural}, which sets scaling exponents for initialization and learning rate with respect to model width $n$. Under $\mu$P, all weights in the model are initialized to have roughly $1/\sqrt{n}$ magnitude (or more precisely $1/\sqrt{fan\_in}$), which implies that features $\Zout$ and their inputs $\Zin$ to have $\Theta_n(1)$ norm at initialization (i.e.  $n^{-1}\|\Zin\|_1= \Theta_n(1)$).

\cref{eq:one_step_signsgd} describes the evolution of features $\Zout$ as we update weights $W$. Ideally, we want both stability ($W_t \Zin$ does not grow in magnitude with $n$) and non-triviality ($W_t \Zin$ does not converge to $0$ with $n$). These conditions are both satisfied when $W_{t+1}\Zin - W_t \Zin = \Theta_n(1)$ element-wise, which implies that the learning rate should scale as $\alpha = \eta n^{-1}$ for some constant $\eta>0$, to compensate the growth in $\|\Zin\|_1$, which is exactly the $\mu$P scaling rule for the learning rate. See \cref{app:add_theory} for more details about the mechanisms of $\mu$P. With this parametrization of the learning rate, we obtain
\begin{align*}
\|W_{t+1}\Zin\|^2_2 = \|W_t\Zin\|_2^2 + \eta^2 n^{-1} \|\Zin\|_1^2
  - 2 \eta n^{-1}\|\Zin\|_1 \times  \langle W_t\Zin, \Sign(d\Zout^t)\rangle.
\end{align*}

We can normalize by $n$ so terms on both sides have $\Theta_n(1)$ magnitude in width $n$,

\begin{align*}
n^{-1}\|W_{t+1}\Zin\|^2_2 = n^{-1}\|W_t\Zin\|_2^2 + \eta^2 n^{-2}\|\Zin\|_1^2
  - 2 \eta n^{-1}\|\Zin\|_1 \times n^{-1} \langle W_t\Zin, \Sign(d\Zout^t)\rangle.
\end{align*}

The term $n^{-1} \langle W_t\Zin, \Sign(d\Zout^t)\rangle$ measures the alignment between the features $\Zout^t = W_t \Zin$ and the ``signed'' gradients $\Sign(d\Zout)$. Intuitively, at the initial training stages, these two terms are roughly independent (as random variables) because of the randomness from the initialization weights. As a result, in those initial training stages, we have 

\begin{equation}\label{eq:roughly_indep}
    \langle W_t \Zin, \Sign(d\Zout)\rangle \approx \bigO(n^{1/2}),
\end{equation}

which yields
\begin{align*}
    n^{-1}\|W_{t+1} \Zin\|_2^2 \approx n^{-1}\|W_t\Zin\|_2^2 + \alpha^2 n^{-2} \|\Zin\|_1^2 + \bigO(n^{-1/2})
\end{align*}
Since $\alpha^2 n^{-2} \|\Zin\|_1^2 = \Theta_n(1)$ is positive and asymptotically non-zero, if the width $n$ is large enough, we should expect the (normalized) feature norm $n^{-1}\|W_{t} \Zin\|_2^2$ to grow initially during training. The next results provides a rigorous description of this phenomenon for linear networks.

\begin{thm}[Feature Norm Growth in Linear Networks (Informal)]\label{thm:growth} 
Assume that the neural network is a linear MLP (see \cref{app:add_theory} for more details). Then, for any $\delta \in (0,1/2)$, under some assumptions stated in \cref{sec:proofs}, there exists a universal constant $\lambda>0$ such that for any $T$ and $\eta$ satisfying $T \leq \lambda \eta^{-1}$, the following holds with probability at least $1-2n^{-1 + 2\delta}$
\begin{equation}
\sup_{1\leq t \leq T}\left|\, n^{-1}\|W_t \Zin\|^2 - \Gamma_t\right| \leq C n^{-\delta},
\end{equation}
where $\Gamma_t = \Gamma_0 + \beta^2 (1 + t(t-1))$,  $\beta = \eta \, n^{-1}\, \|\Zin\|_1$, and $\Gamma_0 = n^{-1}\|W_0 \Zin\|^2$.
In other words, when the width $n$ is large enough, $n^{-1}\|W_t \Zin\|^2$ exhibits quasi-quadratic growth at initial training stages.
\end{thm}

\cref{thm:growth} characterizes the growth in feature norms $n^{-1}\|W_t \Zin\|^2$ as training progresses. The proof is provided in \cref{sec:proofs}. In this case, $n^{-1}\|W_t \Zin\|^2$ grows in a quasi-quadratic pattern, which becomes perfectly quadratic when $n \to \infty$. This is the most important takeway from this result: this phenomenon is associated with large width. With more realistic models, we expect the growth property to hold, but not necessarily with the quadratic form. See next section for empirical results.

\begin{wrapfigure}{r}{0.3\textwidth}
\vspace{-3em}
  \begin{center}
    \includegraphics[width=0.29\textwidth]{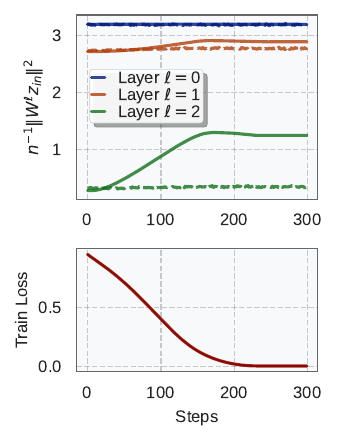}
  \end{center}
  \caption{\small{Evolution of feature norms during training for the linear network described in \cref{subsec:toy_exps}. We train the model for $300$ steps with Adam. Feature norms for different layers exhibit differential growth patterns as we train the model. We shifted the curves corresponding to different layers for better visualization.}}
  \label{fig:toy_model}
  \vspace{-4em}
\end{wrapfigure}
\subsection{Evolution of Feature Norms}\label{subsec:toy_exps}
Consider a three layers linear neural network given by $
f(x) = W_2 W_1 W_0 x,
$
where $x \in \reals^d$, $W_0 \in \reals^{n \times d}$, $W_1 \in \reals^{n \times n}$, and $W_2 \in \reals^{1\times n}$. The training data consist of $N=1000$ datapoints of dimension $d$ generated from a linear model $y = \omega ^\top x + \varepsilon$ with $\varepsilon \sim \normal(0,0.025)$, $\omega_i \sim d^{-1} \normal(0, 1)$,  and $x$ are generated randomly as standard Gaussian random variables. We use $n=d=100$ in our experiments and train the model with Adam. See \cref{sec:experimental_setup} for results with SignSGD and more details about the experimental setup.

\Cref{fig:toy_model} shows the growth in feature norms for the three modules (corresponding to the three layers in this case) as we train the model. We include a baseline (dashed lines) which shows the norms $\|W \ZinT\|$ where $\ZinT$ is a random Gaussian vector with iid coordinates, normalized such that $\|\ZinT\| = \|\Zin\|$ (see next section for an intuitive explanation of this baseline). The baseline does not show any significant growth over the course of the training which further confirms that feature norms grow as a result of increasing alignment between module weights and module inputs, and not simply as a result of an increase in weight norms.

\paragraph{Most of the growth occurs early in training.} Interestingly, most of the growth in feature norms occurs in the first $T=200$ steps, which also correlates with the most significant drop in training loss. After $T=200$, feature norms remain roughly stable until convergence. This suggest that the norm growth is associated with an initial phase where significant feature learning occurs, and remains roughly unchanged after that initial growth phase. Intuitively, as we train the network, the dot product between $\Zout$ and $S(d\Zout)$ (\cref{eq:roughly_indep}) grows from $\bigO(n^{1/2})$ to roughly $\Theta_1(n)$ (in absolute value) and therefore the argument behind the feature norm growth as explained in the discussion above no longer holds later in training. As a result, the growth plateaus after some number of steps $T$.

\paragraph{Different growth levels for different modules.} Although we use the same learning rate for all modules, the norm growth in the input layer ($n^{-1}\|W_0 z_{in_0}\|^2$) is much less significant than that observed in the second layer ($n^{-1}\|W_2 z_{in_2}\|^2$). To understand this difference, we should take into account that when training all modules (layers in this case), the inputs to $W_1$ and $W_2$ change with training. The update in feature are given by
\begin{align*}
W_{t+1}\Zin^{t+1} = W_t\Zin^t - \eta \,n^{-1} \times \|\Zin^t\|_1 \, \Sign(d\Zout^t) + W_{t+1} \Delta \Zin^{t+1},
\end{align*}
where $\Delta \Zin^{t+1} = \Zin^{t+1} - \Zin^t$ is the input change after one step. Under $\mu$P scaling rule for the learning rate ($\eta \, n^{-1}$), the magnitude of $\|\Zin^t\|_1$ remains $\Theta_n(1)$ for all $t$, however the constant in $\Theta_n(1)$ naturally depends on the layer. Additionally, the term $ W_{t+1} \Delta \Zin^{t+1}$ introduces more complex update dynamics, and contribute in a non-trivial way to the change in the feature norms. See \cite{nam2024visualisingfeaturelearningdeep} for a more detailed discussion on how feature learning changes from one layer to another. Both of these aspects lead to uneven growth in the feature norms for different layers.

\paragraph{Different growth levels for different inputs/tasks.} In the setup of \cref{thm:growth}, we considered a batch size of $1$, which results in feature updates of the form  
$
W_{t+1}\Zin = W_t\Zin - \eta \times n^{-1}\|\Zin\|_1 \, \Sign(d\Zout),
$
and we saw that with $1/\sqrt{n}$ initialization, we have  $n^{-1}\|\Zin\|_1 = \Theta_n(1)$. In the realistic setting of batch training, feature updates for an input $x' \in \reals^d$ are given by 
\begin{align*}
W_{t+1}\Zin(x') &= W_t \Zin(x') - \eta \,n^{-1}\times  \Sign\left(\frac{1}{|B|} \sum_{x \in B} d\Zout(x) \otimes \Zin(x)\right) \Zin(x'),
\end{align*}
Therefore, we can no longer directly expand the sign function and obtain the $\|\Zin\|_1$ term that leads to the $\Theta_n(1)$ term.  In this case, we need a strong correlation between $\Sign\left(\frac{1}{|B|} \sum_{x \in B} d\Zout(x) \otimes \Zin(x)\right)$ and $\Zin(x')$ to obtain the same effect. This translates to whether the datapoint $x'$ has some similarity with the batch used for the update. As a result, we should expect to see higher scores for datapoints that are similar to the training dataset, and lower scores for significantly different datapoints.

$\star\star$ \underline{Intuition for our methodology}$\star\star$: The two aspects above (different growth levels for different modules/datasets) provide a compelling approach to measure alignment between \emph{modules} and \emph{datasets}. In the next section, we refine this notion of alignment and use it to create a method for module type selection for LoRA finetuning.

\section{Methodology: Normalized Feature Norms as Alignment Scores}\label{sec:methodology}
Several alignment measures exist in the literature. For instance, \citet{baratin2021implicitregularizationneuralfeature} introduced the centered tangent kernel alignment as a measure of how well aligned each layer is with the task, and \citet{lou2022featurelearning} provided a theoretical analysis of such alignment. \citet{he2024understandingminimisingoutlierfeatures} studied the emergence of large feature norms in the network as a result of different training configurations. Our work introduces a new alignement measure based on the feature norm analysis from the previous section.

Given a pretrained model, and some finetuning dataset $\data$, we calculate feature norms for all modules on the task $\data$ by averaging across a subset of $\data$. This provides information on module alignment with the finetuning dataset $\data$. However, note that the score naturally depends on the norm of $W$ and $\Zin$. In order to capture only the alignment, we need to normalize the feature norm by a baseline feature norm for the same layer. See the discussion after \cref{def:nfn} for an intuitive explanation of this normalization.
\begin{definition}[Normalized Feature Norm (NFN)]\label{def:nfn}
Given a pretrained model, a module with weight $W$ in this model, and an input $x$, we define the Normalized Feature Norm as 
$$
\mathrm{NFN}(W, x) = \frac{\|W \Zin(x)\|}{\|W \ZinT(x)\|} ,
$$
where $\ZinT(x)$ is a vector of the same dimension and norm of $\Zin(x)$, with i.i.d coordinates distributed as centered Gaussian random variables.
\end{definition}

By incorporating the random baseline $\|W \ZinT(x)\|$, NFN score removes the dependence on the norm of $\Zin$ and the matrix norm of $W$. The intuition is simple: with $\ZinT$, we should not expect any alignment with $W$, and therefore that should act as baseline score. More precisely, with the randomized input $\ZinT$, we have
\begin{align*}
W_{t+1}\ZinT &= W_{t}\ZinT - \eta \times \Sign(d\Zout \otimes \Zin) \ZinT \\
&= W_t\ZinT - \eta \times \Sign(\Zin)^\top \ZinT \, \Sign(d\Zout).
\end{align*}
As a result, we obtain $
n^{-1}\|W_{t+1} \ZinT\|_2^2 = n^{-1} \|W_t \ZinT\|_2^2 + \bigO(n^{-1/2}),
$
which implies that no significant growth in $n^{-1}\|W \ZinT\|^2$ should be observed as we train the model, provided that model width is large $n \gg 1$). This is empirically demonstrated in \cref{fig:toy_model} in the previous section. For the NFN scores, intuitively, when the module is well aligned with the data, we expect to see scores NFN$>1$, while the NFN score should be $\approx 1$ when alignment is not significant. 

\begin{figure}[t]
    \centering
    \includegraphics[width=1.1\linewidth]{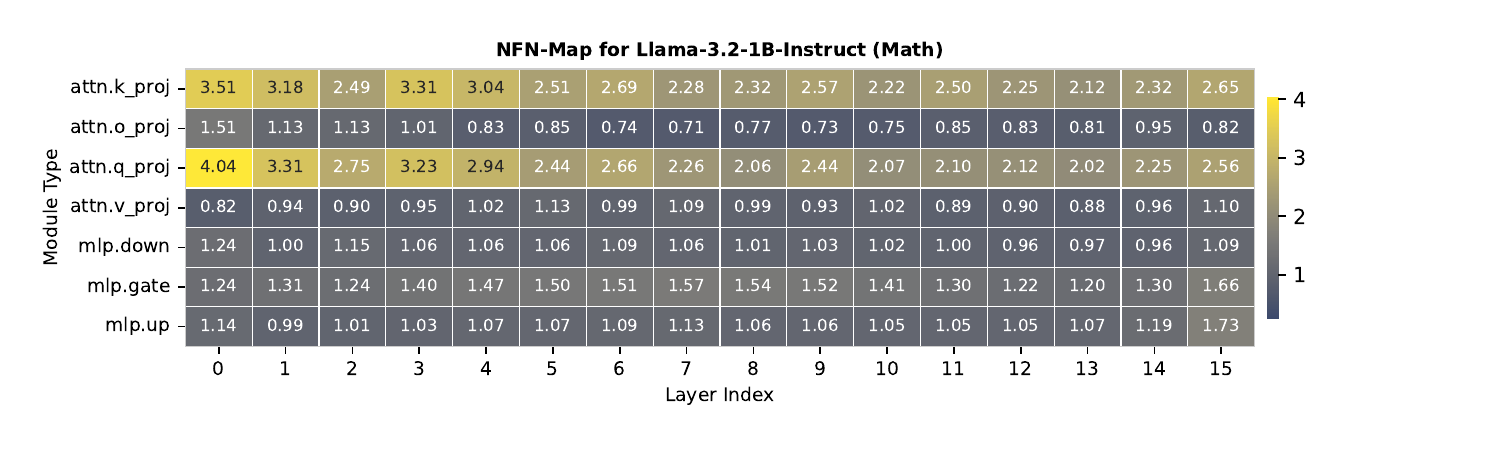}
    \caption{\small{NFN-map for LLama-3.2-1B-Instruct on Math dataset (GSM8K). See \cref{sec:experimental_setup} for NFN-maps of other models.}}
    \label{fig:nfn_map_llama3.2_instruct}
\end{figure}
Under some assumptions on $W$ and $\Zin$, we can prove that when the width is large enough, the NFN score can be approximated by $\textrm{NFN}(W,x)\approx \left\|\frac{W \Zin(x)}{\|W\|_F \|\Zin(x)\|} \right\|$ where $\|W\|_F = \sqrt{\sum_{ij} W_{ij}^2}$ is the Frobenius norm of $W$. This approximation shows that division by $\|W \ZinT \|$ essentially normalizes $W$ and $\Zin$, although not in the standard way where the matrix norm is the operator norm. While both forms are cheap to calculate NFN scores, we prefer the form in \cref{def:nfn} for ease of interpretation.

From this analysis, we can now introduce \method, a method that leverages NFN scores to identify which modules should be prioritized for LoRA finetuning. Our method is described below.
\begin{tabbedbox}{\method~-- \small{Module Type Selection}}
\small{\underline{Inputs}: model $\mathcal{M}$, finetuning dataset $\data$.\\
\textbf{Step1 (Scores)}: calculate $\textrm{NFN}(W, \data) \gets |\data|^{-1} \sum_{x \in \data} \|W\Zin(x)\|^2/\|W\ZinT(x)\|^2$ for all $W$.\\
\textbf{Step2 (Aggregation)}: Calculate $\textrm{NFN}(\textrm{T}, \data) = |N_{T}|^{-1} \sum_{W \in T} \textrm{NFN}(W, \data)$ for all module types $T \in \{$Query, Key, Value, OutProj, GateProj, UpProj, DownProj$\}$.\\
\textbf{Step3 (Insertion)}: Insert LoRA in module types with the lowest NFN scores.}\label{desc:method}
\end{tabbedbox}

We can also think of the reverse \method~method where instead of choosing module types with the lowest scores, we choose the ones with the highest scores. The intuition behind such choice would be that the module types with the highest scores are the ``most'' important for the task. We call this method \method$^{-1}$ and we evaluate its performance in \cref{sec:exps}.

\begin{figure}[t]
    \centering
    \begin{subfigure}{0.47\textwidth}
        \centering
        \includegraphics[width=\linewidth]{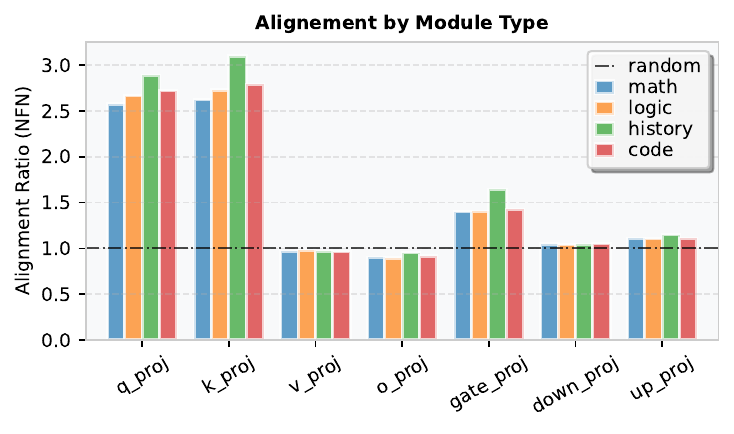}
        \caption{\textbf{Llama3.2-1B-Instruct}}
    \end{subfigure}
    \hfill
    \begin{subfigure}{0.47\textwidth}
        \centering
        \includegraphics[width=\linewidth]{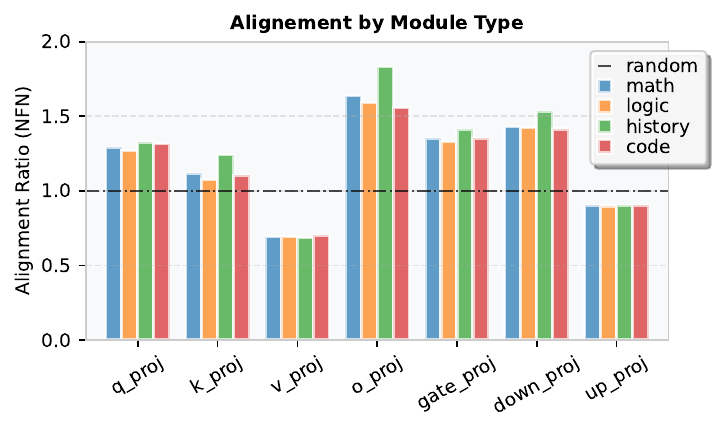}
        \caption{\textbf{Gemma3-1B-Instruct}}
    \end{subfigure}
    \par\bigskip
    \begin{subfigure}{0.47\textwidth}
        \centering
        \includegraphics[width=\linewidth]{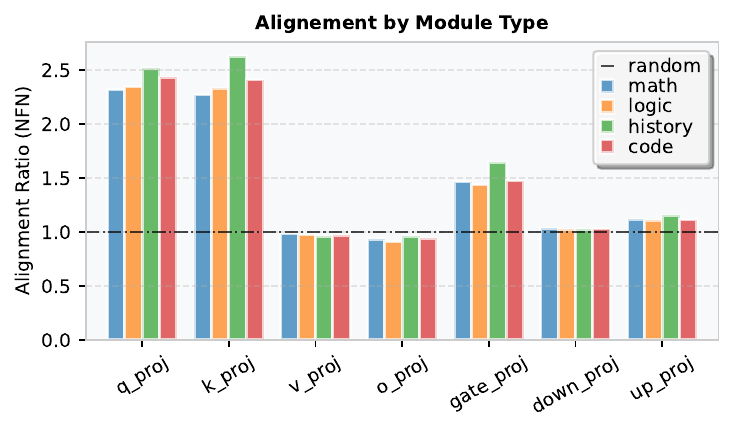}
        \caption{\textbf{Llama3.2-3B-Instruct}}
    \end{subfigure}
    \hfill
    \begin{subfigure}{0.47\textwidth}
        \centering
        \includegraphics[width=\linewidth]{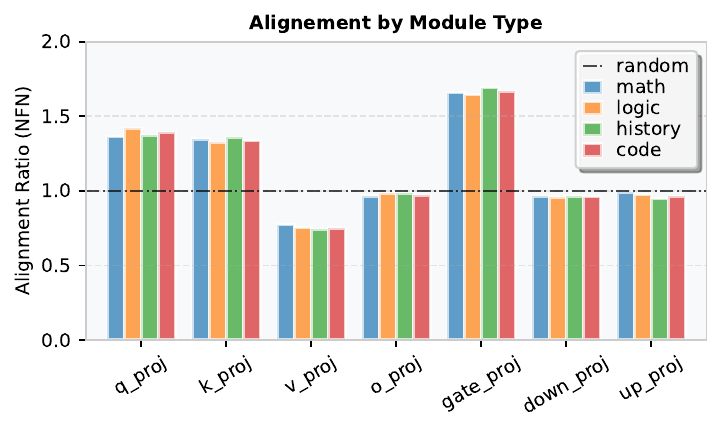}
        \caption{\textbf{Qwen3-1.7B-Instruct}}
    \end{subfigure}
    
    \caption{\small{NFN scores aggregated by module type for different models. The scores are for different datasets (math, code, history, and logic).}}
    \label{fig:nfn_module_type}
\end{figure}

\Cref{fig:nfn_map_llama3.2_instruct} shows NFN scores for LLama-3.2-1B-Instruct for 4 tasks: \emph{math} (GSM8K, \cite{cobbe2021gsm8k}), \emph{code} (HumanEval, \cite{chen2021humaneval}), \emph{history} (MMLU high school european history, \cite{hendrycks2021measuringmassivemultitasklanguage}), and \emph{logic} (MMLU logical fallacies). The NFN-map provides the most granular level of scoring and shows the NFN scores by module. We can see that key and query attention modules are most aligned with the task in this case, while MLP modules are less aligned, suggesting the need for adaptation in those modules. To see this by module type, we aggregate by averaging over all modules of the same type (step 2 in \method) and show the results in the \cref{fig:nfn_module_type} for different models. We observe significant variability of NFN scores accross model, module type, and dataset. For Llama3.2-1B,  module types with the highest scores (Query, Key) average around 2-3X the baseline ($\approx 1$), and the lowest scores (Value, Gate, Down, Up) hovering around the baseline score of 1. In this case, \method~indicates that adaptation should be focused on the (value, gate, down, up) modules rather than the attention query and key matrices. Note that this coincides with the recommendation of empirical work by \cite{he2021towards} for Llama models but is contradictory to the recommendations of \cite{hu2022lora} to finetune mainly attention modules. 

Qwen3-1.7B shows high alignment in Query, Key, and Gate modules, with lower alignment for other MLP modules, and a surprisingly low score for the Value module ($\approx 0.75$). This indicates that the Value modules in Qwen3-1.7 are ``negatively'' aligned with with all datasets, suggesting that inputs to the Value modules are aligned with the smallest singular directions of the Value weight matrices. The same pattern can be observed in Gemma3-1B, and we currently do not have an explanation for this phenomenon. In \cref{sec:experimental_setup}, we provide NFN scores for additional Qwen, Gemma, and Llama models.

\paragraph{NFN scores are sensitive to tasks.} The alignment scores differ between tasks. For instance, model weights show larger alignment with \texttt{history} compared to \texttt{math}, suggesting that their training data consisted more of sequences similar to general natural language than math related tokens, which is expected. However, note that all tasks share some ``base'' alignment level given by the general magnitude of the NFN score for each module type. This is a more fundamental phenomenon that is independent of the task, and is  related to some basic level of feature learning that is required for token processing. \footnote{The mechanisms of feature learning in deep neural networks are still largely misunderstood. Quantitative approaches such as \cite{nam2024visualisingfeaturelearningdeep} offer some insights, but are far from being comprehensive.}

\paragraph{NFN scores consistent across different model sizes.} In \cref{fig:nfn_module_type} (a) and (c), we show NFN scores for two model sizes of Llama3.2, 1B and 3B. The ranking of module types based on NFN scores is roughly the same for both models, suggesting consistency of NFN scores across different model sizes. Intuitively, having similar NFN score patterns suggests similar pretraining and post-training processes for these models, which is expected for models of the same family (Llama3.2 in this case).

\begin{figure}[t]
    \centering
    \includegraphics[width=0.43\linewidth]{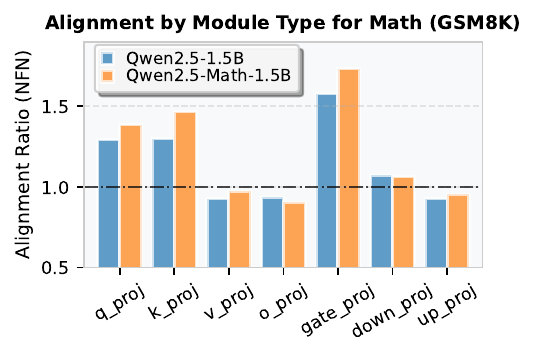}
    \includegraphics[width=0.445\linewidth]{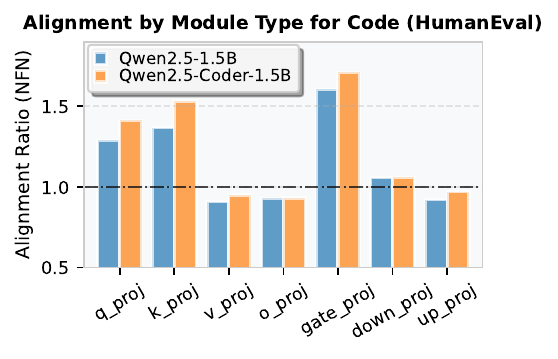}
    \caption{\small{Module types NFN scores for general and specialized Qwen2.5 models. Specialized models (math, code) are finetuned on task-specific data. Scores are higher with the specialized models.}}
    \label{fig:nfn_specialized}
\end{figure}

\paragraph{Specialized Models show higher NFN scores.} In \cref{fig:nfn_specialized}, we compare NFN scores for instruction-tuned and more specialized version of the same model Qwen2.5-1.5B for math/code tasks. As expected, the specialized models show higher NFN scores overall which further confirms that NFN scores, while cheap to calculate, can be a reliable measure of module-data alignment.  

\paragraph{Compute cost of \method.} To obtain the results in \cref{fig:nfn_map_llama3.2_instruct}, we used a single forward pass with batch size 200, with a maximum sequence length of 256. NFN scores are calculated using the \texttt{register\_hook} functionality of PyTorch. In summary, the computational cost of our method is roughly the same as a single batch forward pass, which makes it especially relevant in resource-constrained environments where LoRA is most useful.

In the next section, we run extensive experiments to show that \method~ consistently enhances final performance at virtually no added cost.

\section{Experiments}\label{sec:exps}
We conduct extensive experiments to verify the benefits of using \method~ in different scenarios. We consider three post-training scenarios: Supervised Finetuning (SFT) for classification, Supervised Finetuning for text generation, and Reinforcement Learning (GRPO, \cite{shao2024deepseekmathpushinglimitsmathematical}), all with LoRA adapters. We report results with Llama \citep{2024llama3herdmodels}, Qwen \citep{qwen3}, and Gemma \citep{gemmateam2025gemma3technicalreport} models across different sizes. Our experiments are as follows:

\begin{enumerate}
    \item SFT for classification: we finetune classifers on ANLI \citep{nie2019adversarial}.
    \item SFT for text generation: we train on MetaMathQA \citep{yu2023metamath} and evaluate the results on GSM8K \citep{cobbe2021gsm8k}.
    \item GRPO: we conduct a study on on the effect of RL on mathematical reasoning, where the model is trained on MetaMathQA and evaluated on GSM8K.
\end{enumerate}

We investigate the effect of different module placement strategies: our method \method (placing LoRA in module types with the lowest NFN scores), \method$^{-1}$  (the inverse of our method, i.e. placing LoRA modules types with the highest NFN scores), Attn (inserting LoRA only in attention modules), MLP (inserting LoRA only in MLP modules), and ALL (inserting LoRA in all module types).

Hereafter, we use the following letters to denote specific modules: Q (Query), K (Key), V (Value), O (Out projection), U (Up projection), G (Gate projection), and D (Down projection).
All module type NFN scores and experimental details are provided in \cref{sec:experimental_setup}.

\subsection{Supervised Finetuning for Classification}
\begin{figure}
    \centering
    \includegraphics[width=0.49\linewidth]{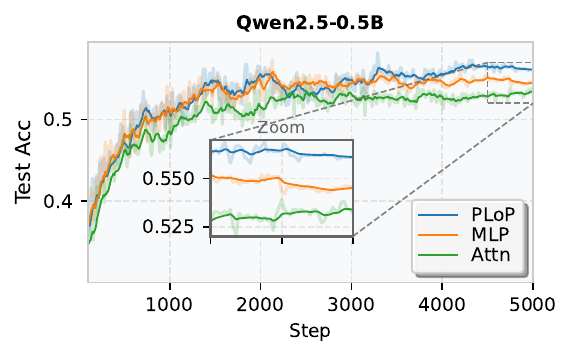}
    \includegraphics[width=0.49\linewidth]{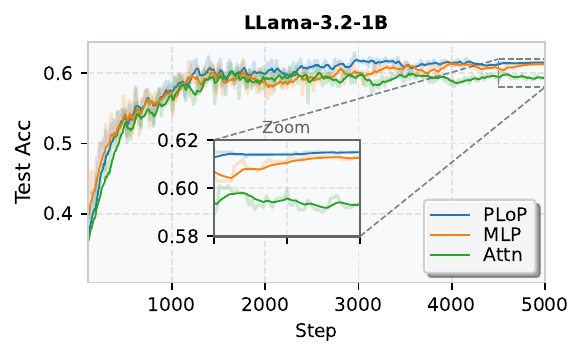}
    \caption{\small{Results of LoRA finetuning on ANLI for different models. We use LoRA rank $r=8$ for MLP strategy and adapt $r$ for \method~and Attn to match number of parameters for fair comparison. All curves are smoothened with EMA($\alpha=0.8$) for better visualization. See \cref{app:sft_setup} for more details about the experimental setup. (\textbf{Left}) For Qwen2.5-0.5B, \method~places adapters in V, O, and Up modules. (\textbf{Right}) For Llama3.2-1B, \method~places adapters in V, O, and Down modules.}}
    \label{fig:classification_anli}
\end{figure}

The Adversarial Natural Language Inference (ANLI) is a language classification task that is considered more challenging compared to similar tasks (e.g. MNLI). Using LoRA with different placement strategies, we finetune pretrained models on ANLI and report results in \cref{fig:classification_anli}. For Qwen2.5-0.5B, we observe a significant difference in performance between \method~and other strategies. For Llama3.2-1B, \method~and MLP yield roughly the same performance, while Attn is significantly worse. Note that for the Llama model, the MLP modules have small NFN scores, comparable to scores of modules selected by \method (V-O-D, see \cref{fig:nfn_module_type}), which could explain why we get similar performance with \method~and MLP.

\subsection{Supervised Finetuning for Text Generation}
Supervised finetuning is an important component of the post-training pipeline as it plays an important role in improving model abilities such as mathematics, coding, and instruction following. Often, finetuning data is high-quality and specifically curated to provide dense signal for the model to acquire specific desirable skills. For challenging tasks such as mathematical reasoning, it is often used to ``prime'' the model for subsequent RL based finetuning which we investigate in the next section. 

The finetuning task we consider is mathematics. We perform finetuning on the MetaMathQA dataset \cite{yu2023metamath}. To evaluate the performance we measure the accuracy on the GSM8k benchmark \cite{cobbe2021gsm8k}. For a training configuration with LoRA rank $r$ we set LoRA $\alpha = 2r$ and optimize using Adam. For each adapter placement we sweep over the learning rate in $\{1, 2, 3, 4, 5\} \times 10^{-4}$ and report the result with the best accuracy. We provide additional experimental details regarding the training and evaluation configuration in Appendix \ref{app:sft_setup}.

\begin{table}[htbp]
  \centering
  \setlength{\tabcolsep}{4pt}
  \renewcommand{\arraystretch}{1.1}
  \scriptsize

  \parbox{\textwidth}{\centering\large\bfseries MetaMathQA $\rightarrow$ GSM8k}
  \vspace{0.1ex}

  \begin{minipage}[t]{0.48\textwidth}
    \centering
    \captionof{table}{\textbf{Qwen3-0.6B}}
    \label{tab:qwen3-0.6b_math}
    \begin{tabularx}{\linewidth}{%
        >{\raggedright\arraybackslash}X
        >{\centering\arraybackslash}p{0.14\linewidth}
        >{\centering\arraybackslash}p{0.14\linewidth}
        >{\centering\arraybackslash}p{0.14\linewidth}
      }
      \rowcolor{headercolor}
      \textcolor{white}{\textbf{Module Types}} & 
      \textcolor{white}{\textbf{\#Params (M)}} & 
      \textcolor{white}{\textbf{Train Loss}} & 
      \textcolor{white}{\textbf{Eval Acc}} \\
      \midrule
      \rowcolor{rowcolor1} \method(D–U–V)  ($r = 76$)        & 21.8 & 0.118 & 63.8\% \\
      \rowcolor{rowcolor1} \method(D–U–V)  ($r = 64$)        & 18.4 & 0.116 & 62.0\% \\
      \rowcolor{rowcolor1} \method$^{-1}$(G–K–Q)  \tiny{($r = 64$)} & 16.5 & 0.119 & 60.6\% \\
      \rowcolor{rowcolor1} MLP(D–G–U) ($r = 64$)             & 22.0 & 0.119 & 63.3\% \\
      \rowcolor{rowcolor1} Attn(K–Q–V) ($r = 64$)             & 12.8 & 0.130 & 58.6\% \\
      \rowcolor{rowcolor1} all ($r = 64$)               & 40.4 & 0.113 & 62.4\% \\
      \bottomrule
    \end{tabularx}
  \end{minipage}
  \begin{minipage}[t]{0.48\textwidth}
    \centering
    \captionof{table}{\textbf{Qwen3-1.7B}}
    \label{tab:qwen3-1.7b_math}
    \begin{tabularx}{\linewidth}{%
        >{\raggedright\arraybackslash}X
        >{\centering\arraybackslash}p{0.14\linewidth}
        >{\centering\arraybackslash}p{0.14\linewidth}
        >{\centering\arraybackslash}p{0.14\linewidth}
      }
      \rowcolor{headercolor}
      \textcolor{white}{\textbf{Module Types}} & 
      \textcolor{white}{\textbf{\#Params (M)}} & 
      \textcolor{white}{\textbf{Train Loss}} & 
      \textcolor{white}{\textbf{Eval Acc}} \\
      \midrule
      \rowcolor{rowcolor1} \method(D–O–V)  ($r = 102$)        & 43.9 & 0.109 & 75.4\% \\
      \rowcolor{rowcolor1} \method(D–O–V)  ($r = 64$)        & 27.5 & 0.111 & 75.2\% \\
      \rowcolor{rowcolor1} \method$^{-1}$(G–K–Q)  \tiny{($r = 64$)} & 27.5 & 0.113 & 74.6\% \\
      \rowcolor{rowcolor1} MLP(D–G–U) ($r = 64$)             & 44.0 & 0.108 & 75.0\% \\
      \rowcolor{rowcolor1} Attn(K–Q–V) ($r = 64$)             & 18.4 & 0.119 & 69.5\% \\
      \rowcolor{rowcolor1} all ($r = 64$)               & 69.7 & 0.105 & 73.9\% \\
      \bottomrule
    \end{tabularx}
  \end{minipage}
\end{table}
We perform finetuning on the Qwen3 models with 0.6B and 1.7B parameters. The results are shown in Tables \ref{tab:qwen3-0.6b_math} and \ref{tab:qwen3-1.7b_math} respectively. In both cases, the placement of adapters in only the attention layers is clearly suboptimal, demonstrating that for challenging tasks such as mathematics, adapting only the attention layers has limited effect. We see that the most competitive placements for both model sizes is to place adapters according to $\method$ or in the MLP layers, even outperforming the placement of adapters in all layers which requires between 1.5-1.8$\times$ the number of trainable parameters. When adjusting for an equal number of trainable parameters, $\method$ produces the best results with a slight edge over the MLP placement. For the 1.7B parameter model $\method$ has a small edge over the MLP placement even with about $60\%$ the number of parameters.

\subsection{Reinforcement Learning for Enhanced Reasoning}
Reinforcement Learning has emerged as a promising approach for test-time scaling. Algorithms such as GRPO work by incentivizing the model to follow a pattern of ``thinking'' before providing the final answer. This implicit approach to reasoning (versus the more explicit approaches such as MCTs \citep{xie2024montecarlotreesearch}) showed very promising results, especially with the impressive performance of DeepSeek-R1 \citep{deepseekai2025deepseekr1incentivizingreasoningcapability}. In this section, we experimented with ``GRPO on a budget'' using LoRA adapters (instead of training the full weights) to enhance mathematical reasoning.  We select 3 module types for LoRA adapter placement using the three approaches stated above. We compare performance both during RL training (columns Rwd/Format and Rwd/Answer) and Eval (GSM8K 8shots prompting Pass@1). Note that because LoRA is a lightweight finetuning method, it would not be enough to induce reasoning in base models with GRPO, especially with a small rank $r$. For this purpose (and in contrast to DeepSeek-R1), we apply GRPO with LoRA to instruction-tuned models instead of base models. For more implementation details, see \cref{sec:experimental_setup}.
\begin{table}[htbp]
    \centering
    \footnotesize
    \caption{GRPO results for \textbf{Qwen3-1.7B} trained on MetaMathQA.}
    \label{tab:qwen3-1.7b}
    \begin{tabular}{>{\raggedright\arraybackslash}p{3.7cm}
                    >{\centering\arraybackslash}p{2.cm}
                    >{\centering\arraybackslash}p{2.cm}
                    >{\centering\arraybackslash}p{2.cm}
                    >{\centering\arraybackslash}p{2.cm}}
    \arrayrulecolor{headercolor}
    \rowcolor{headercolor}
    \textcolor{white}{\textbf{Module Types}} & 
    \textcolor{white}{\textbf{\#Params (M)}} & 
    \textcolor{white}{\textbf{Rwd/Format}} & 
    \textcolor{white}{\textbf{Rwd/Answer}} 
    & 
    \textcolor{white}{\textbf{Eval/GSM8K}} \\
    \midrule
    \rowcolor{rowcolor1}
    No RL & -- & -- & -- & 65.50\% \\
    \rowcolor{rowcolor1}
    Attn(Q-K-V) ($r=16$) & 4.58 & 1.89 & 0.91 & 71.49\% \\
    \rowcolor{rowcolor1}
    Attn(Q-K-V) ($r=25$) & 7.17 & 1.97 & 0.98 & 72.13\% \\
    \rowcolor{rowcolor1}
    MLP(U-G-D) ($r=16$) & 11.01 & 2.57 & 1.28 & 73.61\% \\
    \rowcolor{rowcolor1}
    \method$^{-1}$(Q-K-G) ($r=16$) & 6.88 & 1.71 & 0.86 & 71.41\% \\
    \rowcolor{rowcolor1}
    \method (V-O-D) ($r=16$) & 6.88 & 2.67 & 1.32 & 74.52\%\\
    \rowcolor{rowcolor1}
    \method (V-O-D) ($r=25$) & 10.75 & 2.75 & 1.32 & 75.03\% \\
    \rowcolor{rowcolor1}
    \bottomrule
    \end{tabular}\label{tab:grpo_qwen3}
\end{table}

With GRPO, we define the think-then-answer pattern in similar way to DeepSeek-R1. The model is rewarded for placing the thinking process in between \texttt{<think>} and \texttt{</think>}, and then giving the answer in between \texttt{<solution>} and \texttt{</solution>}. This is encoded in the format reward function (Rwd/Format). The correctness of the solution is rewarded as well (Rwd/Answer). We track this reward as we train the model with GRPO and show the final results (at convergence). 

\cref{tab:grpo_qwen3} shows the results of GRPO on the Qwen3-1.7B trained on MetaMathQA. \method~yields better performance overall both during training (rewards) and evaluation (GSM8K). Compared to Attn, \method~performs better even when equalizing the number of trainable parameters (Attn($r=25$) vs \method($r=16$)). Interestingly, \method~performs better than MLP placement strategy even when using the same rank $r=16$, in which case we have 6.88M trainable parameters with \method~and 11.01M parameters with MLP. For the same number of parameters, the performance is further improved with \method($r=25$).

We experimented with Gemma3-1B as well, and we found that \method~outperforms other alternatives, although the score on GSM8K is low due the inherent limitations of Gemma3-1B in mathematical reasoning. See \cref{sec:experimental_setup} for more details.

\section{How-To-Use for Practitioners}
The code for \method~is available at: \url{https://github.com/soufiane001/plop}, and full code for GRPO and SFT experiments will be released soon.

To compute the NFN scores, practitioners can use our provided code with minimal setup. After installing dependencies, simply run the following command to compute NFN scores for the \texttt{meta-llama/Llama-3.2-1B-Instruct} model on a given dataset (in this case, math):
\begin{center}
\begin{minipage}{0.95\linewidth}
\begin{verbatim}
python main.py --model meta-llama/Llama-3.2-1B-Instruct \
    --dataset math \
    --batchsize 8 \
    --nbsamples 100 \
    --seqlen 256 \
    --aggregation type \
    --output_dir results/
\end{verbatim}
\end{minipage}
\end{center}
This command will automatically download the specified model and dataset, process the data in batches, and compute the NFN scores for all modules, then aggregate them by module type. Of course, the dataset can be customized as needed. 

An example of the aggregated NFN scores by module type is shown below:
\begin{center}
\begin{minipage}{0.3\linewidth}
\begin{verbatim}
===========================
 NFN Scores by Module Type
===========================
 q_proj: 2.58
 k_proj: 2.63
 v_proj: 0.97
 o_proj: 0.90
 gate_proj: 1.40
 down_proj: 1.05
 up_proj: 1.11
===========================
\end{verbatim}
\end{minipage}
\end{center}
In this case, we should add LoRA adapters in module types \texttt{v\_proj}, \texttt{o\_proj}, and \texttt{down\_proj}.

This workflow enables straightforward targeted LoRA finetuning across different architectures and datasets.

\section{Discussion and Limitations}\label{sec:discussion}
In this paper, we introduced \method, an intuitive module type selection method, designed specifically for LoRA fine-tuning and based on NFN scores -- a notion of module-data alignment supported by theory. \method~meets the computational criteria needed for efficient LoRA finetuning as articulated in the introduction: \method's lightweight nature makes it particularly valuable in resource-constrained environments where LoRA is most beneficial. \method~is based on the NFN-map, which enables more granular selection beyond module types. However, we deliberately focused on module type selection as it represents the most widely adopted aggregation approach among practitioners, avoiding the additional implementation complexities of more fine-grained selection. While we explored using \method~for layer-level selection by inserting LoRA into target layers with low NFN scores, we encountered inconsistent results and have reserved this question for future research.

\section{Acknowledgment}
We gratefully acknowledge partial support from NSF grants DMS-2209975 and DMS-2413265, NSF grant 2023505 on Collaborative Research: Foundations of Data Science Institute (FODSI), the NSF and the Simons Foundation for the Collaboration on the Theoretical Foundations of Deep Learning through awards DMS-2031883 and 814639,  NSF grant MC2378 to the Institute for Artificial CyberThreat Intelligence and OperatioN (ACTION), and NSF ACCESS grant CIS250193.

\bibliography{refs}
\newpage
\appendix

\section{Additional theoretical details}
\label{app:add_theory}

\subsection{\texorpdfstring{Infinite-width analysis and $\mu$P}{Infinite-width analysis and muP}}
Scaling remains the main paradigm to improve performance of language model (see e.g. \cite{hoffmann2022training}). This includes model capacity which can be increased via width (embedding dimension) or depth (number of layers) or both, training data, number of training steps etc. In our theoretical analysis in \cref{sec:theory}, we mentioned the infinite-width $n \to \infty$ and how our results hold in this limit. This is motivated by the fact that most state-of-the-art language and vision models have large width.

As the width $n$ grows, most hyperparameters in the model such as the initialization and the learning should be adapted to avoid numerical instabilities and ensure efficient learning. For instance, the initialization variance should scale as $1/n$ to prevent arbitrarily large pre-activations as we increase model width $n$ (e.g. He init \cite{he2015delvingdeeprectifierssurpassing}). To derive such scaling rules, a principled approach consist of analyzing statistical properties of key quantities in the model (e.g. pre-activations) as $n$ grows and then adjust the initialization, the learning rate, and the architecture itself to achieve desirable properties in the limit $n \to \infty$ \cite{hayou19activation, yang2019scaling}.

In this context, \citet{yang2022featurelearninginfinitewidthneural} introduces the Maximal Update Parameterization (or $\mu$P), a set of scaling rules for the initialization scheme, the learning rate, and the network architecture that ensure stability and maximal feature learning in the infinite width limit. Stability is defined by $Y_l^i = \Theta(1)$ for all $l$ and $i$ where the asymptotic notation `$\Theta(.)$' is with respect to width $n$ (see next paragraph for a formal definition), and feature learning is defined by $\Delta Y_l = \Theta(1)$, where $\Delta$ refers to the feature update after taking a gradient step. $\mu$P guarantees that these two conditions are satisfied at any training step $t$. Roughly speaking, $\mu$P specifies that hidden weights should be initialized with $\Theta(n^{-1/2})$ random weights, and weight updates should be of order $\Theta(n^{-1})$. Input weights should be initialized $\Theta(1)$ and the weights update should be $\Theta(1)$ as well. While the output weights should be initialized $\Theta(n^{-1})$ and updated with $\Theta(n^{-1})$. These rules ensure both stability and feature learning in the infinite-width limit, in contrast to standard parameterization (exploding features if the learning rate is well tuned), and kernel parameterizations (e.g. Neural Tangent Kernel parameterization where $\Delta Y_l = \Theta(n^{-1/2})$, i.e. no feature learning in the limit). 
\section[Proof of Theorem~\ref{thm:growth}]{Proof of \texorpdfstring{\cref{thm:growth}}{Theorem~\ref{thm:growth}}}\label{sec:proofs}
In this section, we provide the full proof for \cref{thm:growth}. Forst, we prove a result on the sign of the derivative of the loss function with respect to $\Zout$, then proceed with the full proof.

\subsection{Constant loss derivative sign in the initial training stage}
Consider a linear network of the form 
\begin{equation}
    f(x) = W_L W_{L-1} \dots W_0 x, \quad x \in \reals^d,
\end{equation}
where $L \geq 1$ is the network depth, $W_{\ell} \in \reals^{n\times n}$ for all $\ell \in \{1, 2, \dots, L-1\}$ are hidden layers parameters,  $W_L \in \reals^{1\times n}$ is the projection layer weight, and $W_0 \in \reals^{n\times d}$ is the input layer weight.

The network is trained with the setup described in \cref{sec:theory}, namely:
\begin{itemize}
    \item Dataset: a single datapoint $(\hat{x}, \hat{y}) \times \reals^d \times \reals$.
    \item Training algorithm: SignSGD with learning rate $\eta n^{-1}$.
    \item Single layer: only a single hidden layer that we denote $W \in \{W_\ell, \ell=1,2,\dots, L-1\}$ is trained, and other layers weights are fixed to their values at initialization. Without loss of generality, assume that the trainable layer is $\ell_0$, i.e. $W = W_{\ell_0}$.
    \item Training loss: quadratic loss given by $\mathcal{L}(W) = 2^{-1} (f_W(\hat{x}) - \hat{y})^2$. 
\end{itemize}

In this setting, the linear network training dynamics become tractable and we can obtain closed-form expressions in steps $t$ and width $n$. Let us use the same notation as in \cref{sec:theory} and denote the input and output of the trainable layer $\Zout = W \Zin$. More precisely, in this case, we can express the network output as $f_W(x) = V^\top \Zout = V^\top W \Zin$, where $\Zin = M x$, with $M = W_{\ell_0 - 1} \dots W_0 \in \reals^{n\times d}$ and $V^\top = W_L \dots W_{\ell_0 + 1} \in \reals^{1\times n}$ are both non-trainable random matrices. In this case, the gradient of the loss with respect to $\Zout$ is given by 

$$
d\Zout = (V^\top W \Zin - y) V.
$$

From now on, we will abuse the notation and use the subscript to denote the training step as well for the matrix $W = W_{\ell_0}$. When we use the notation $W_t$, it should interpreted as $W_{\ell_0, t}$. Taking one step with SignSGD yields

$$
W_{t+1} = W_t - \eta\, n^{-1} \|\Zin\|_1 \chi_t \Sign(V), 
$$
where $\Sign(.) = sign(.)$ and $\chi_t = \Sign(V^\top W_t \Zin - \hat{y})$.

Next, we prove a result that will be useful in the proof of \Cref{thm:growth}. More precisely, we show that under mild assumptions, there exists a first initial training phase in which the sign of the loss function on the training datapoint does not change. The number of steps in this phase is bounded by $\eta^{-1}$ up to some constant factor. Naturally, since we initialize with random variables, it should be expected that such result could only hold with high probability. 

\begin{thm}[Constant loss derivative sign in the initial training phase]\label{thm:constant_derivative_sign}
We assume that the weights $W_0, W_1, \dots, W_L$ are initialized such that the following holds:
\begin{itemize}
    \item  $|\Zin^i| \in [\munderbar{Z}, \bar{Z}]$ for all $i \in [1:n]$, where $\munderbar{Z}, \bar{Z} >0$ are constants independent of $n$. 
    \item Mean-field Init: $\E[V_i] = 0$ and $\textrm{Var}(V_i) = n^{-2}$ (e.g. uniform distribution on $[-n^{-1}, n^{-1}]$).
\end{itemize}
Further assume that $y \in [\munderbar{Z}, \bar{Z}]$. \footnote{This can satisfied with a simple adjustment of the constants $\munderbar{Z}, \bar{Z}$.}

Then, for any $\delta \in (0,1)$, and $T\leq \eta^{-1} \left( n^{-\delta} + \frac{\bar{Z}}{\munderbar{Z}^2}\right)$, we have with probability at least $1-n^{-1+\delta}$, 
$$
\forall t\leq T, \chi_t = \chi_0.
$$
\end{thm}
The assumption on the weight initialization is mild and is satisfied by some standard initialized schemes, such as uniform init with $n^{-1}$ variance for the hidden weights, $d^{-1}$ variance for the input weights, and $n^{-2}$ variance for the projection weights. The proof of \cref{thm:constant_derivative_sign} relies on standard concentration results.

\begin{proof}
Recall the definition of $\chi_t$

$$
\chi_t = \Sign(V^\top W_t \Zin - \hat{y}).
$$

We have the following from above, 

$$
W_t \Zin = W_{t-1} \Zin - \eta n^{-1} \|\Zin\|_1 \chi_{t-1}\, \Sign(V),
$$
and therefore, 
$$
V^\top W_t \Zin = V^\top W_{t-1} \Zin - \eta n^{-1} \|\Zin\|_1 \|V\|_1 \chi_{t-1}\,
$$

which implies that
\[
V^\top W_t \Zin = V^\top W_0 \Zin - \eta n^{-1} \|\Zin\|_1 \|V\|_1 \left[ \sum_{j=0}^{t-1} \chi_j \right].
\]

\textbf{Bounding $V^\top W_0 \Zin$:}

With Chebyshev's inequality we have:
\[
\mathbb{P} \left( |V^\top W_0 \Zin | \geq \bar{Z} n^{-\delta} \right) 
< \frac{\mathrm{Var}(V^\top W_0 \Zin)}{(\bar{Z} n^{-\delta})^2}
\]

where 
\[
\mathrm{Var}(V^\top W_0 \Zin) = \frac{1}{n} \mathrm{Var}(W_0^i{}^\top \Zin) = \frac{1}{n} \cdot \frac{\|\Zin\|^2}{n} \leq n^{-1} \bar{Z}^2
\]

As a result, we obtain:
\[
\mathbb{P} \left( |V^\top W_0 \Zin| \geq \bar{Z} n^{-\delta} \right) < n^{-1 + 2\delta}
\]

If  $V^\top W_0 \Zin - \hat{y} < 0$, we have
\[
V^\top W_t \Zin - \hat{y} \leq V^\top W_0 \Zin - \hat{y} + \eta n^{-1} \|\Zin\|_1 T
\]
\[
\leq V^\top W_0 \Zin - \hat{y} + \eta \bar{Z} T
\]

With probability at least $1 - n^{-1 + 2\delta}$, we have
\[
V^\top W_t \Zin - \hat{y} \leq \bar{Z} n^{-\delta} - \hat{y} + \eta \bar{Z} T
\]

Therefore, we have
\[
T \leq \eta^{-1} \left( \frac{\hat{y}}{\bar{Z}}- n^{-\delta} \right)
\Rightarrow \forall t \leq T,\quad \chi_t = \chi_0 = -1.
\]

If $V^\top W_0 \Zin - \hat{y} > 0$,  asymptotically this implies that $-\hat{y} >0$ (assuming $|\hat{y}| = \Theta_n(1)$). Similarly, we obtain with probability at least $1-n^{-1+2\delta}$,

\[
V^\top W_t \Zin - \hat{y} \geq  V^\top W_0 \Zin - \hat{y} - \eta \bar{Z} T,
\]
\[
 \geq  - \bar{Z} n^{-\delta} - \hat{y} - \eta \bar{Z} T,
\]

and therefore, we have that 
\[
T \leq \eta^{-1} \left( \frac{- \hat{y}}{\bar{Z}}- n^{-\delta} \right)
\Rightarrow \forall t \leq T,\quad \chi_t = \chi_0 = 1.
\]

In summary, we have the following: Let $\delta \in (0,1)$. Then, with $T \leq \eta^{-1} \left( \frac{|\hat{y}|}{\bar{Z}} - n^{-\delta} \right)$, we have for all $t \leq T$, $\chi_t = \chi_0$.

\end{proof}

The assumptions in \cref{thm:constant_derivative_sign} can be alleviated to include more generalization initialization schemes, such as non-clipped Gaussian initialization. However, this will require additional control on the asymptotics of $\|\Zin\|$, $\|\Zin\|_1$, and $V$. The result remains the same however.

\subsection{\texorpdfstring{Proof of \Cref{thm:growth}}{Proof of Theorem~\ref{thm:growth}}}

\textbf{Theorem \ref{thm:growth}} .[Feature Norm Growth in Linear Networks]\\
\emph{Assume that the neural network is linear. Then, for any $\delta \in (0,1/2)$, under the assumptions on the initialization stated in \cref{thm:constant_derivative_sign}, there exists a universal constant $\lambda>0$ such that for any $T$ and $\eta$ such that $T \leq \lambda \eta^{-1}$, the following holds with probability at least $1-2n^{-1 + 2\delta}$
\begin{equation}
\sup_{1\leq t \leq T}\left|\, n^{-1}\|W_t \Zin\|^2 - \Gamma_t\right| \leq C n^{-\delta},
\end{equation}
where $\Gamma_t = \Gamma_0 + \beta^2 (1+t(t-1))$,  $\beta = \eta \, n^{-1}\, \|\Zin\|_1$, and $\Gamma_0 = n^{-1}\|W_0 \Zin\|^2$.
In other words, $n^{-1}\|W_t \Zin\|^2$ exhibits quasi-quadratic growth at early training phase, when the width is sufficiently large.}\\

\begin{proof}
Recall the update with SignSGD 
$$
W_{t+1} = W_t - \eta\, n^{-1} \chi_t\,  \Sign(V) \otimes \Zin, 
$$

where $\Sign(.) = sign(.)$ and $\chi_t = \Sign(V^\top W_t \Zin - \hat{y})$.

Denoting $\alpha_t = \langle W_t \Zin, \Sign(V) \rangle$, we obtain 

$$
\alpha_{t+1} = \alpha_t - \beta \chi_t \times n = \alpha_0 - \beta \, n \sum_{j=0}^t \chi_t.
$$

Therefore, 
\begin{align*}
    \|W_{t+1} \Zin\|^2_2 &= \|W_{t} \Zin\|_2^2 + \eta^2 n^{-2} \|\Zin\|_1^2 \times n - 2 \eta n^{-1} \|\Zin\|_1 \chi_t \times \alpha_t\\
    &= \|W_{t} \Zin\|_2^2 + \beta^2 \times n - 2 \beta \chi_t \times \alpha_t.
\end{align*}

Let $\delta \in (0,1/2)$. From \cref{thm:constant_derivative_sign}, it is straightforward that there exists a constant $\lambda>0$ such that for any $T > 1 $ and $\eta$ such that $T\leq \lambda \eta^{-1}$, with probability at least $1-n^{-1+\delta}$, we have for all $t\leq T, \chi_t = \chi_0$. In this case, for $t \leq T$, we have $\chi_t \times \alpha_t = \chi_t \times \alpha_0 - \beta n \sum_{j=0}^{t-1} \chi_t \times \chi_{j} = \chi_t \times \alpha_0 - \beta n \times  t$. 

Therefore,
$$
n^{-1}\|W_{t+1} \Zin\|^2 = n^{-1}\|W_{t} \Zin\|^2  + \beta^2 + 2 \beta^2 t - 2 \beta  \chi_t n^{-1} \alpha_0.
$$

Using Chebyshev's inequality, we can easily show that for any $\delta \in (0,1)$, with probability at least $1- n^{-1+2\delta}$, we have 
$$
|\chi_t n^{-1} \alpha_0| \leq \bar{Z} n^{-\delta},
$$
which yields that with at least the same probability we have 

$$
|n^{-1}\|W_t \Zin\|^2 - \Gamma_t| \leq |n^{-1}\|W_{t-1} \Zin\|^2 - \Gamma_{t-1}| + 2 \beta \bar{Z} n^{-\delta}, 
$$

where we define the sequence $\Gamma_{t+1} = \Gamma_t + \beta^2 ( 1 + 2t)$, with $\Gamma_0 = n^{-1}\|W_0 \Zin\|^2$. Then, it is straightforward that for all $t\leq T$ 
$$
|n^{-1}\|W_t \Zin\|^2 - \Gamma_t| \leq 2 \beta \bar{Z} T n^{-\delta}.
$$

With union bound, this occurs with probability at least $1-2n^{-1+\delta}$.
\end{proof}

Note that the probability bound can be significantly improved by considering sub-gaussian concentration bounds instead of Chebyshev's inequality. Since our aim in this paper is mainly methodological, we do not include it here.

\section{Additional Experimental Details}\label{sec:experimental_setup}

\subsection{Experimental Setup for the linear network}
The linear network is given by 

$$
f(x) = W_2 W_1 W_0 x,
$$

where $x \in \reals^d$, $W_0 \in \reals^{n\times d}$, $W_1\in \reals^{n\times n}$, and $W_2 \in \reals^{1,n}$.

\paragraph{Dimensions.} We use $d=n= 100$ in our experiments.

\paragraph{Training Data.} We generate a random vector $w\in \reals^d$ with iid coordinates $w_i \sim d^{-1/2} \normal(0,1)$ and fix it for the next step. Then, we generate $N=1000$ samples from the following distribution:
\begin{itemize}
    \item $x \sim \reals^d$ random vector with iid coordinates $x_i \sim \normal(0,1)$
    \item $y = w^\top x + \epsilon$, where $\epsilon \sim \normal(0,0.025)$
\end{itemize}

\paragraph{Training.} We use Adam algorithm for training, and train the model for $T=300$ steps with full batch.

\subsection{Experimental Setup for SFT (Classification)}
For ANLI experiments, we use the following training configuration
\begin{itemize}
    \item Training datasets: ANLI
    \item Training algorithm: AdamW, no warmup, linear schedule, dropout (0.1).
    \item Max sequence length 256.
    \item LoRA $\alpha = 2r$
    \item Precision: bf16. 
\end{itemize}

We use $r=8$ for MLP placement stratgy, and adapt $r$ to match param count for other placement strategies. Specifically:
\begin{itemize}
    \item Qwen3.5-0.5B: MLP ($r=8$), Attn ($r=36$), \method($r=17$)
    \item Llama3.2-1B: MLP ($r=8$), Attn ($r=27$), \method($r=15$)
\end{itemize}

\subsection{Experimental Setup for SFT (Text Generation)}\label{app:sft_setup}
For the SFT experiments we use the following training configuration
\begin{itemize}
    \item Training dataset: MetaMathQA
    \item Training algorithm: Adam
    \begin{itemize}
        \item epochs: 2
        \item warmup: 0.1 fraction
        \item schedule: cosine
        \item no dropout
    \end{itemize}
    \item Max sequence length 1024.
    \item LoRA $\alpha = 2r$
    \item Precision: bf16. 
\end{itemize}

For evaluation on GSM8k we use the script \texttt{evaluate\_chat\_gsm8k.py} in the official \href{https://github.com/QwenLM}{QwenLM repo}. We evaluate with 8-shot examples using the Qwen chat template. We apply a strict match for evaluating the accuracy and allow 512 generation tokens.

\subsection{Experimental Setup for GRPO}
For GRPO, we use the following config: 

\begin{itemize}
    \item Training Dataset: Subset of MetaMathQA (50k samples).
    \item Training Alg: AdamW with warmup (0.1) and weight decay (0.01) and cosine schedule. We use LR 4e-6 for all training runs. We found this to be a good LR in our experiments. Unlike exps in SFT, we could not run sweeps over LR for GRPO due to limited computational resources and the high cost of GRPO runs, but we expect LR tuning to further improve the results.
    \item Precision: bf16
    \item Number of generations: 8
    \item Maximum generation length: 512
    \item Batch size: 64 (16 with 4 steps for gradient accumulation)
    \item LoRA dropout: 0.05
    \item Rewards: A combination of reward functions (correctness, format)
    \item Hardware: 2xGH200 GPUs
\end{itemize}

We use custom eval script for GSM8K (using the chat template of each model).

\subsection{Additional Empirical Results}

\subsection{GRPO results for Gemma3}

\begin{table}[htbp]
    \centering
    \caption{GRPO results for Gemma3-1B trained on MetamathQA \citep{yu2023metamath}.}
    \label{tab:qwen3-1.7b_app}
    \begin{tabular}{>{\raggedright\arraybackslash}p{3.7cm}
                    >{\centering\arraybackslash}p{2.cm}
                    >{\centering\arraybackslash}p{2.cm}
                    >{\centering\arraybackslash}p{2.cm}
                    >{\centering\arraybackslash}p{2.cm}}
    \arrayrulecolor{headercolor}
    \rowcolor{headercolor}
    \textcolor{white}{\textbf{Module Types}} & 
    \textcolor{white}{\textbf{Rwd/Format}} & 
    \textcolor{white}{\textbf{Rwd/Answer}} 
    & 
    \textcolor{white}{\textbf{Eval/GSM8K}} \\
    \midrule
    \rowcolor{rowcolor1}
    No RL &  -- & -- & 29.10\% \\
    \rowcolor{rowcolor1}
    Attn (Q-K-V) ($r=16$)  & 2.16 & 0.89 & 30.05\% \\
    \rowcolor{rowcolor1}
    MLP (U-D-G) ($r=16$) & 2.11 & 0.88 & 29.81\% \\
    \rowcolor{rowcolor1}
    \method$^{-1}$ (O-G-D) ($r=16$)  & 1.91 & 0.86 & 28.05\% \\
    \rowcolor{rowcolor1}
    \method (K-V-U) ($r=16$) & 2.36 & 0.92 & 30.52\%\\
    \rowcolor{rowcolor1}
    \bottomrule
    \end{tabular}\label{tab:grpo_gemma3}
\end{table}

Interestingly, for Gemma3 1B, we found that most of the RL rewards was accumulated in forms of format reward (placing the thinking process between \texttt{<think>} and \texttt{</think>} and the solution between \texttt{<answer>} and \texttt{</answer>}). This is reflected in \cref{tab:grpo_gemma3}. However, for eval on GSM8K, we found that accuracy after GRPO didn't change significantly which is probably due the fact that Gemma3-1B is weak on such tasks. In such cases, LoRA is probably not suitable, and full finetuning is needed to enhance reasoning capabilities.

\subsection{Additional NFN-Maps}

\subsubsection{Qwen3-1.7B-Instruct}

\begin{figure}[ht]
    \centering
    \includegraphics[width=1.2\linewidth]{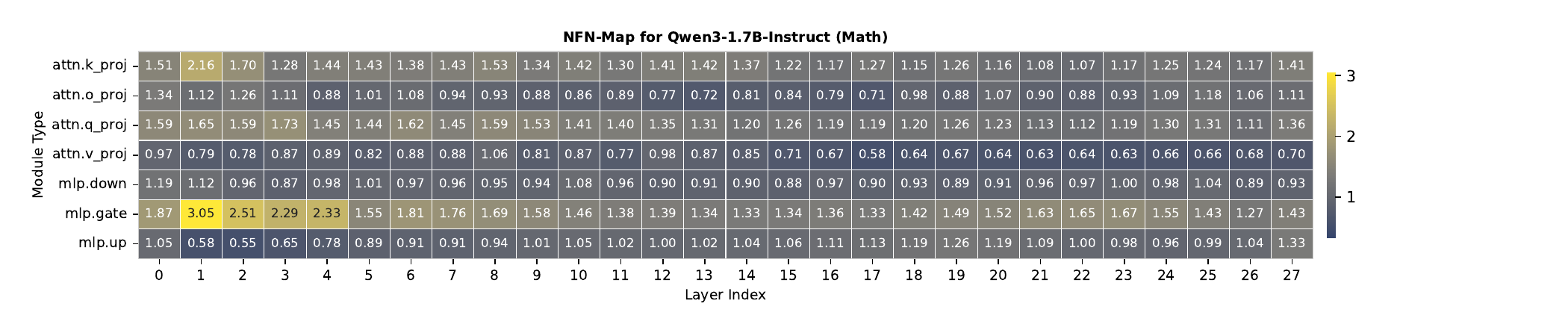}
    \includegraphics[width=.7\linewidth]{figures/metrics/comparison_type_Qwen3-1.7B-Instruct.pdf}
    \caption{NFN scores for Qwen3-1.7B}
    \label{fig:qwen3_type_nfn}
\end{figure}

\subsubsection{Qwen2.5-3B-Instruct}

\begin{figure}[H]
    \centering
    \includegraphics[width=1.2\linewidth]{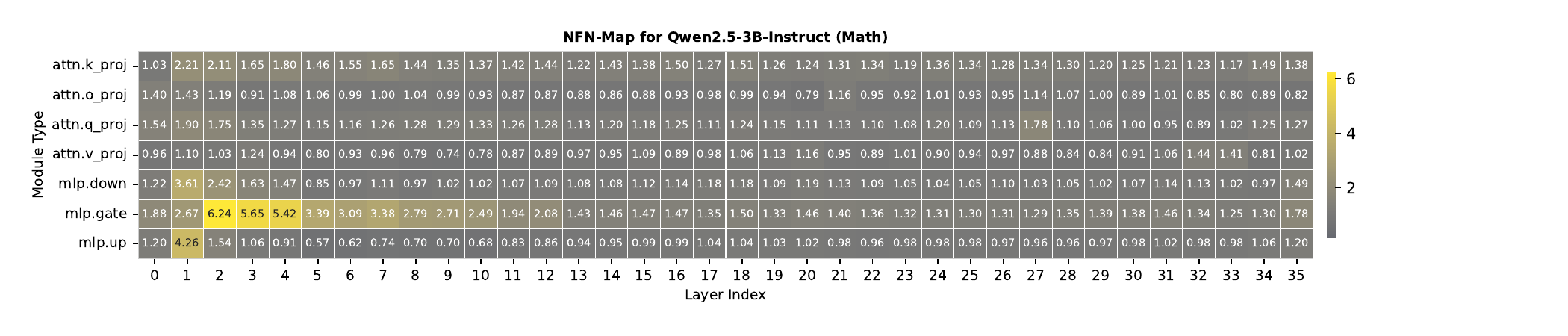}
    \includegraphics[width=.7\linewidth]{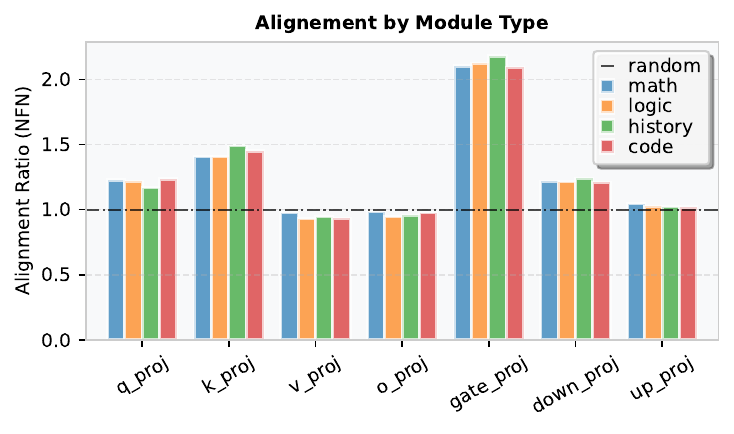}
    \caption{NFN scores for Qwen2.5-3B}
    \label{fig:qwen2.5_3b_type_nfn}
\end{figure}

\subsubsection{Qwen2.5-1.5B-Instruct}

\begin{figure}[H]
    \centering
    \includegraphics[width=1.2\linewidth]{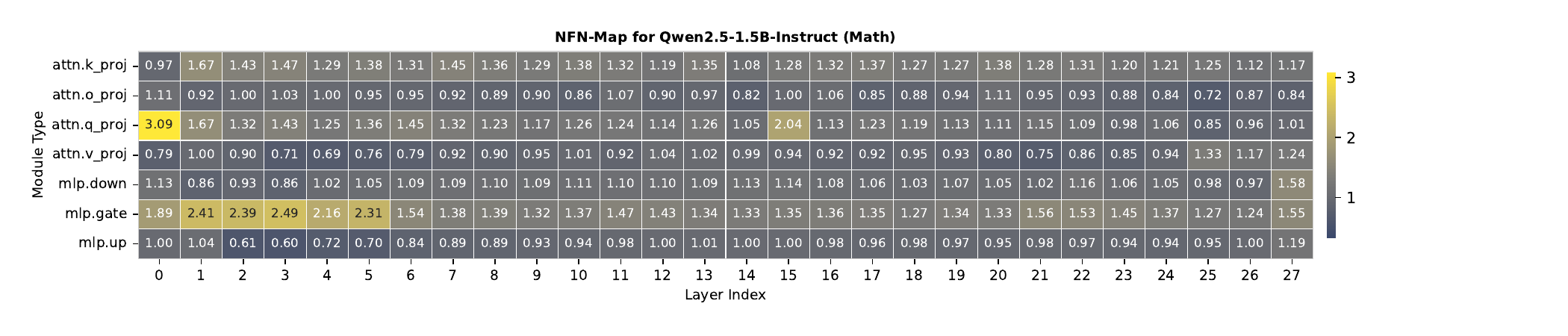}
    \includegraphics[width=.7\linewidth]{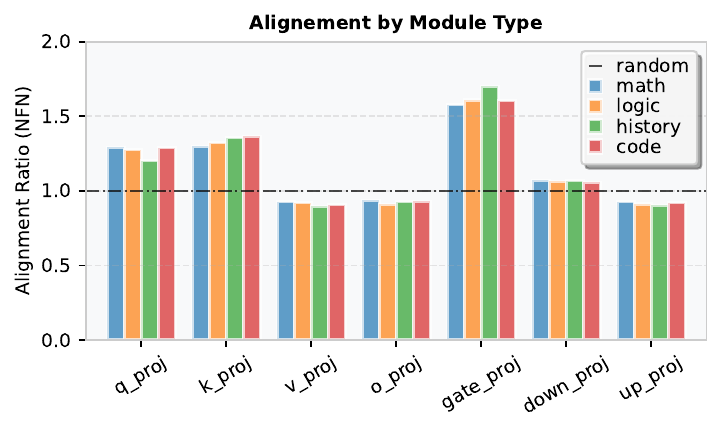}
    \caption{NFN scores for Qwen2.5-1.5B}
    \label{fig:qwen2.5_type_nfn}
\end{figure}

\subsubsection{Qwen2.5-1.5B-Coder-Instruct}

\begin{figure}[H]
    \centering
    \includegraphics[width=1.2\linewidth]{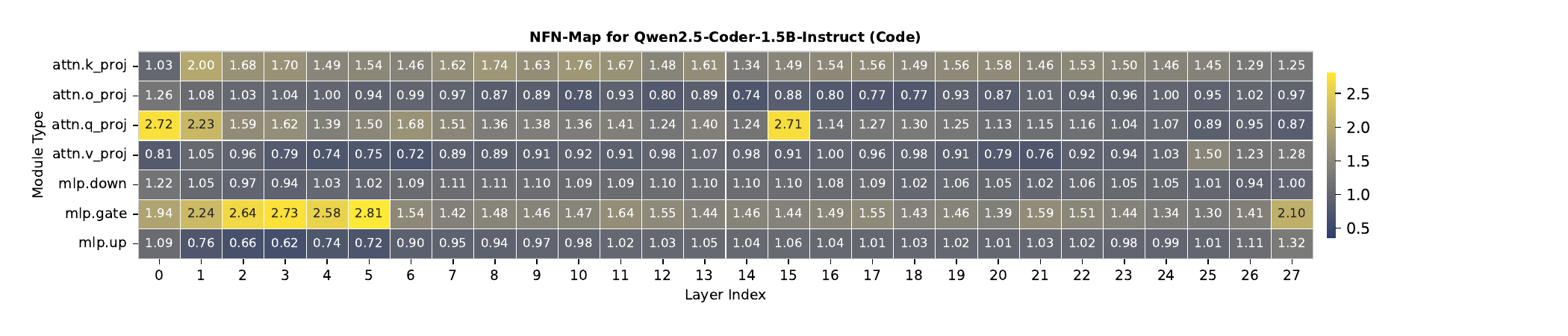}
    \includegraphics[width=.7\linewidth]{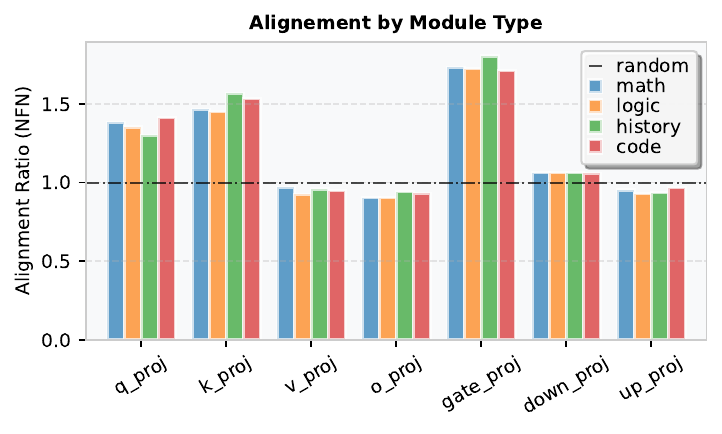}
    \caption{NFN scores for Qwen2.5-1.5B-Coder}
    \label{fig:qwen2.5_coder_type_nfn}
\end{figure}

\subsubsection{Gemma3-1B-Instruct}

\begin{figure}[H]
    \centering
    \includegraphics[width=1.2\linewidth]{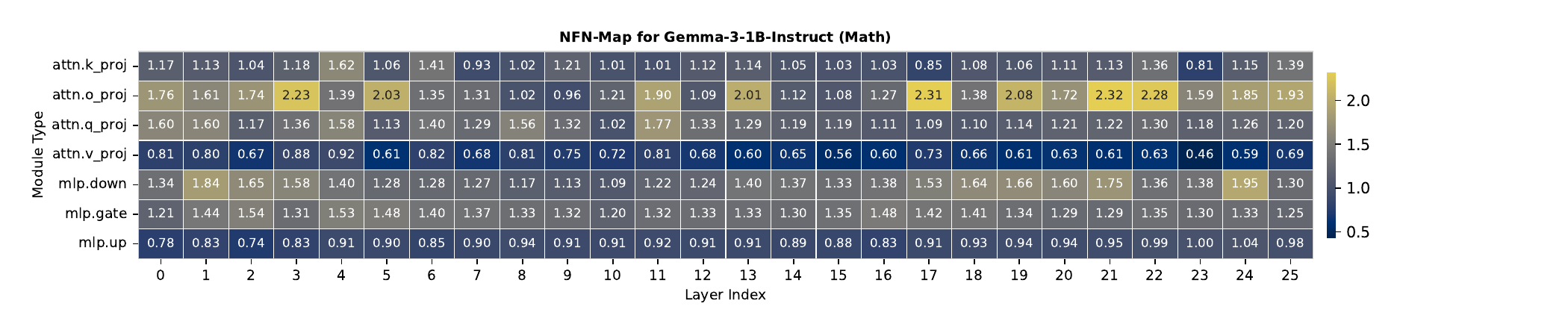}
    \includegraphics[width=.7\linewidth]{figures/metrics/comparison_type_Gemma-3-1B-Instruct.pdf}
    \caption{NFN scores for Gemma3-1B-Instruct}
    \label{fig:gemma3_1b_type_nfn}
\end{figure}

\end{document}